\documentclass{article}
\usepackage[utf8]{inputenc}
\usepackage[dvipsnames]{xcolor}
\usepackage{parskip}
\usepackage{comment}
\usepackage{amsfonts}
\usepackage{amsmath}
\usepackage{multicol}

\usepackage[margin=1.25in]{geometry}

\title{Sequential Models in the Synthetic Data Vault}
\author{
 Kevin Zhang\\
  \texttt{kevz@mit.edu}\\
  MIT
  \and
  Kalyan Veeramachaneni\\
  \texttt{kalyanv@mit.edu} \\
  MIT \\
    \and
  Neha Patki\\
  \texttt{neha@sdv.dev}\\
  \and
  The Synthetic Data Vault team\footnote{All developers who have contributed to the open source SDV project}
}

\date{June 2022}

\usepackage{biblatex}
\usepackage{algorithm}
\usepackage{algpseudocode}

\usepackage{graphicx}
\usepackage{multirow}

\usepackage{appendix}

\begin{document}
\maketitle
\begin{abstract}

The goal of this paper is to describe a system for generating synthetic sequential data within the Synthetic data vault. To achieve this, we present the Sequential model currently in SDV, an end-to-end framework that builds a generative model for multi-sequence, real-world data. This includes a novel neural network-based machine learning model, conditional probabilistic auto-regressive (CPAR) model. The overall system and the model is available in the open source Synthetic Data Vault (SDV) library \footnote{https://github.com/sdv-dev/DeepEcho} \footnote{https://github.com/sdv-dev/SDV}, along with a variety of other models for different synthetic data needs.

After building the Sequential SDV, we used it to generate synthetic data and compared its quality against an existing, non-sequential generative adversarial network based model called CTGAN. To compare the sequential synthetic data against its real counterpart, we invented a new metric called Multi-Sequence Aggregate Similarity (MSAS). We used it to conclude that our Sequential SDV model learns higher level patterns than non-sequential models without any trade-offs in synthetic data quality.

\end{abstract}

\section{Introduction}

Synthetic data is machine-generated data that is created specially with the goal of mimicking the format and mathematical properties of real data. Its applications range from protecting the privacy of real data to creating enhanced, augmented datasets for data science. A few years back we created an open source ecosystem called the Synthetic Data Vault (SDV), with a goal to be the most comprehensive and trusted set of approaches for creating synthetic data. To that end, the open source SDV library offers a variety of models suited for different usages ranging from the original, multi-table SDV model \cite{sdv} to \texttt{CTGAN}, a popular, GAN-based generative model \cite{ctgan}. SDV also provides a benchmarking system called \texttt{SDGym}, a set of metrics to evaluate synthetic data via a library called \texttt{SDMetrics} and a set \textit{reversible data transforms} (called \texttt{RDT}) that allow several data types to be converted to numeric formats such that they can be modeled using generative models. 

With our abstractions and feedback from community of researchers, our ability to create new models outpaced our ability to present them in a mathematically rigorous way. Researchers and users have consistently requested to have such presentation. This paper is an attempt to describe the first sequential model in the SDV. In this first version we describe an end-to-end system for generating synthetic data that is sequential. In the next two subsections we motivate the need to be able to model sequential data and generate synthetic versions of it and our contributions through this paper. 

In section~\ref{seqdata} we describe the properties of the sequential data that we focus on. In section~\ref{framework} we present the end to end framework, followed by the CPAR model in section~\ref{cpar}. We then present the experimental setup and evaluation framework in section~\ref{eval}, metrics to evaluate sequential synthetic data in section~\ref{metrics}, and results in section~\ref{results}. We conclude with our discussion in section~\ref{discussion}.  



\subsection{Why Sequential Data?}

With over 300K downloads of the SDV project, we frequently encounter users who want model data that is stored in a single table format. A significant portion of this data is sequential, meaning that the data points occur in a particular order, for example:

\begin{itemize}
  \item a time series of measurements taken at regular intervals, such as in health care monitoring signals or meteorology,
  \item streams of data points representing irregular events, such as clicks on a website or transactions in a market,
  \item generally ordered data, such as delivery stops in a transportation service
\end{itemize}

As our examples show, sequential data is ubiquitous across a wide range of industries. Furthermore this type of data requires special consideration for machine learning because there is an order between the rows of each sequence.

\subsection{Contributions}

In this paper, we present a new model for creating synthetic sequential datasets for real-world data. Our contributions include:

\begin{enumerate}
  \item \textbf{Sequential SDV}, an end-to-end framework for modeling and creating synthetic sequential data. This can be applied on a variety of enterprise datasets that are not necessarily cleaned and formatted. The implementation is available in the open source SDV library \cite{sdv-dev}.
  \item \textbf{CPAR Model}, an innovation on existing Probabilistic Auto Regressive neural network models that allows conditional inputs and uses a custom loss function.
  \item \textbf{MSAS}, an algorithm for measuring the similarity between real and synthetic sequential data.
\end{enumerate}

The rest of this paper is organized as follows: In Section 2, we define the properties of sequential data and present the overall Sequential SDV framework. In Section 3, we describe the algorithmic and mathematical details of the neural network-based CPAR model. In Section 4, we evaluate our sequential model and compare it to non-sequential models.


\section{Sequential SDV}\label{seqdata}
The Sequential SDV framework is designed to handle a variety of real world datasets. In this section, we'll start by defining the properties of sequential data we have observed in the real world and end by describing our framework for handling them.

\subsection{Definitions \& Properties of Sequential Data}
In this section, we define and break down different aspects of sequential data that we have found in real world datasets and provide a mathematical notation for it.

\subsubsection{A Mixed Data Types Sequence}

We define a \emph{sequence} as a dataset with inter-row dependencies. We denote it as $S = S_0, S_1, ...$, where each value $S_t$ represents a row in the $t^{th}$ position. Each row, $S_t$, is dependent on all the rows that come before it in order: $S_0, S_1, ..., S_{t-1}$.

\begin{figure}[H]
    \centering
    \includegraphics[width=0.65\linewidth]{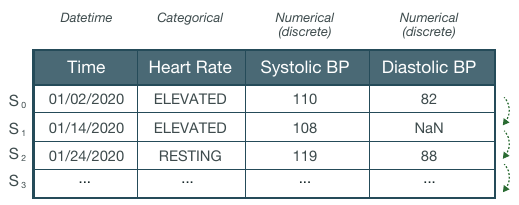}
    \caption{This example shows some fictitious health-related data. Each row $S_t$ represents a set of measurements for the heart rate, systolic and diastolic blood pressure. They are indexed by the timestamp column. Each row is dependent on the ones before it, as shown by the dotted arrows. And each is a mix of datetime, categorical and numerical data types.}
    \label{fig:single-seq-example}
\end{figure}

A sequence may optionally include a separate column, such as a timestamp, that reinforces that order. We call this a \emph{sequence index}. The index values may occur at regular or irregular intervals. The sequence may also contain a mix of data types, for example datetime, numeric, categorical, boolean, etc. Moreover, some of the values may be missing. Figure \ref{fig:single-seq-example} illustrates these properties of a sequence.

\subsubsection{A Multi-Sequence}
We have often found that real-world datasets can contain multiple sequences in a single table. In a \emph{multi-sequence} dataset, different sequences exist independently of each other in the same table. Only the set rows that belong to the same sequence have an inter-row dependency.

In this paper, we denote each sequence with a different superscript, as $S^{(0)}, S^{(1)}, etc$. Note that there is no ordering between different sequences. However the rows belonging to the same sequence continue to have an specific order, denoted as $S^{(i)}_0, S^{(i)}_1, etc$.

\begin{figure}[H]
    \centering
    \includegraphics[width=0.75\linewidth]{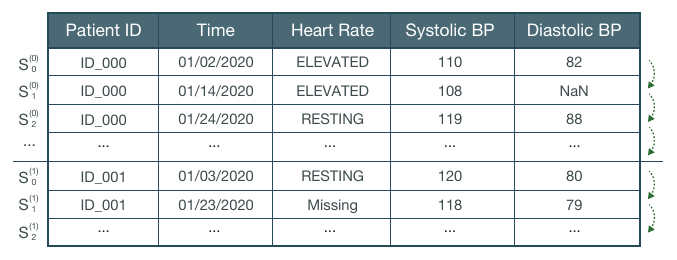}
    \caption{This multi-sequence data contains different sequences as defined by the sequence key, Patient ID. Patient 0's data is $S^{(0)}$, Patient 1's data is $S^{(1)}$, etc. The data for Patient 0 is independent from Patient 1. However, within a single patient's sequence, $S^{(i)}_0, S^{(i)}_1, ...$, there are inter-row dependencies.}
    \label{fig:multi-seq-example}
\end{figure}

How do we know which rows belong to which sequences? Multi-sequence data must contain at least one additional column with this information. We call this the \emph{sequence key}. This is illustrated in Figure \ref{fig:multi-seq-example}.

The sequences may be of different lengths and the intervals of measurements may not match up between them.  

\subsubsection{Contextual Information}
Finally, we have often found that real-world datasets can contain information that does not vary within a sequence. We refer to this data as \emph{contextual information}, denoting it as a constant, $C$. Similar to before, we can use the notation $C^{(0)}, C^{(1)}, etc$. to denote the different contexts for the different sequences.

\begin{figure}[H]
    \centering
    \includegraphics[width=0.85\linewidth]{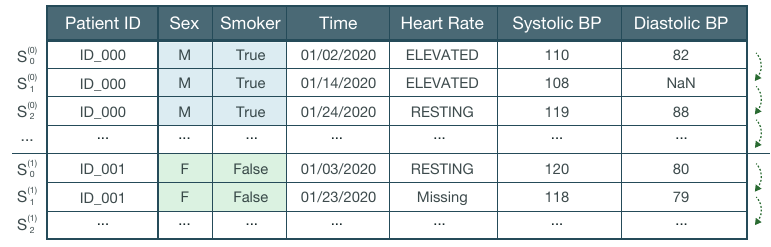}
    \caption{This example shows a multi-sequence dataset with context: the Sex and Smoker columns describes attributes about the patient that do not change over the course of the sequence. If $S^{(0)}$ refers to the sequence for Patient 0, then $C^{(0)}$ refers to its unchanging context: $(M, True)$.}
    \label{fig:context-sequence-example}
\end{figure}

\textbf{Summary.} To create synthetic sequential data for real world datasets, it's important to consider the different properties that the sequential data may have.
\begin{itemize}
    \item A mix of data types include numeric, categorical, datetime, etc, including some missing values.
    \item The possibility that multiple sequences can be present in the same table and the possibility that they may have different lengths.
    \item The possibility that there may be context data that doesn't change over the course of the sequence. Each sequence has a different context.
\end{itemize}

\subsection{Sequential SDV: A Framework for Real World Datasets}\label{framework}
\begin{figure}[H]
    \centering
    \includegraphics[width=0.75\linewidth]{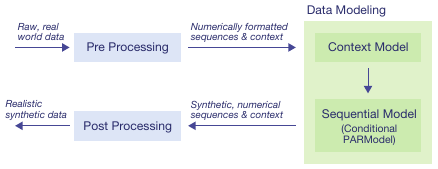}
    \caption{The major components of Sequential SDV are the pre and post processing steps along with the data modeling step. We have broken the modeling into two distinct parts: The context model, for modeling the unchanging contexts, and the sequential model for modeling the sequences.}
    \label{fig:sequential-sdv-summary}
\end{figure}

By design, the Sequential SDV  framework is able to handle mixed type, multi sequence and contextual information. Figure \ref{fig:sequential-sdv-summary} shows the summary of the major components of our framework.

\subsubsection{Pre and Post Processing}
Processing the data is a requirement before and after the data modeling. Before modeling, we need to convert raw data to formats that can be modeled by mathematical and statistical models. In most cases, this is numerical values. After modeling, we need to reverse the changes so that the synthetic data looks like the real data. In this section, we'll discuss the preprocessing step in detail, noting that the postprocessing is simply the reverse of this.

There are 2 types of preprocessing we perform:
\begin{enumerate}
    \item Numerical: This process involves converting all the raw data, which includes categorical values, datetimes, missing values, etc. into numerical data that is ideal for modeling using generative machine learning.
    \item Normalization: This process involves reformatting the sequential data to prepare it for our particular model.
\end{enumerate}

\textbf{Numerical.}
Optimizing the data for machine learning depends on the type of data:

\emph{Missing Data.} If there is any missing data, we replace it with the average of the column. We also create a new column that stores when the original value was missing or not (0 if not missing, 1 if missing)

\emph{Categorical Data.} Unordered, categorical data is represented as a vector of one hot encoded values. If there are $N$ different categories, the vector length will be $N$.

\emph{Continuous Numerical Data.} This data is typically is ready for machine learning as-is. But to optimize the model, we apply a z-score transformation: If $\mu$ is the mean of the column and $\sigma$ is its standard deviation, we transform each value $x$ using the following formula:

\[ Z(x) = \frac{x - \mu}{\sigma} \]

\emph{Discrete Numerical Data.} This type of data goes through a similar optimization. We apply a min-max normalization. If $min$ is the lowest value in the column and $max$ is the highest, then we transform each value $x$ using the following formula: 

\[ N(x) = \frac{x - min}{max- min}\]

\textbf{Normalization.} Our model requires a specific format in order to run. The reformatting applies to any context columns and multi-sequence data:

\begin{enumerate}
    \item If there are any unchanging context columns, $C^{(i)}$, they should all be pulled out in a separate table. We call this table the \emph{context table} and it is no longer sequential.
    \item If there are multiple sequences in the original table, we separate them out into their own tables.
    \item For each sequence, denote the $start$ and $stop$ by adding 2 new columns and 2 new rows.
\end{enumerate}

\begin{figure}[H]
    \centering
    \includegraphics[width=0.95\linewidth]{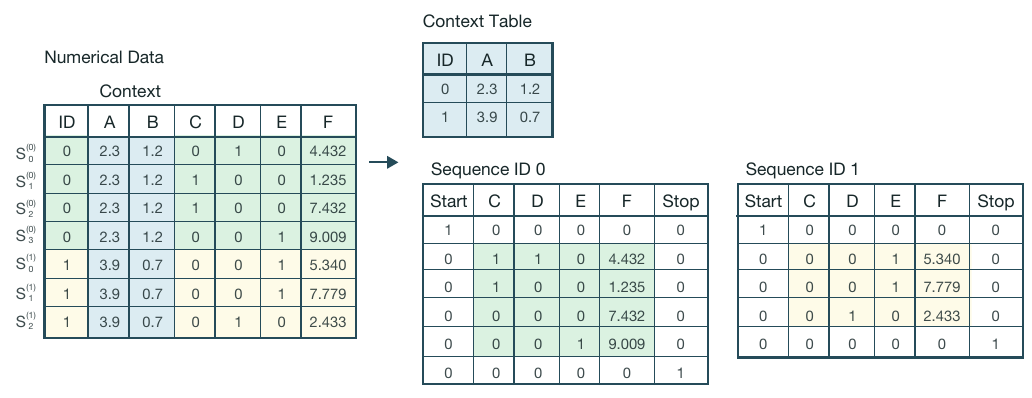}
    \caption{The numerical table at the left has unchanging context columns, A and B shown in blue, as well as two separate sequences, 0 and 1 shown in green and yellow respectively. We first pull out all the context values into its own \textit{context table} (right, blue). Then, we separate out the sequences (right, green and yellow). We add 2 new rows and columns to encode the start and stop for a sequence, using a binary encoding scheme. }
    \label{fig:sequential-processing}
\end{figure}

The normalization steps are illustrated in Figure \ref{fig:sequential-processing}.

Now, we are ready for generative modeling.

\subsubsection{Modeling}
A crucial step in the Sequential SDV framework is to separate out the modeling for the unchanging context versus the sequences.

\begin{enumerate}
    \item We first apply a Gaussian Copula model \cite{sdv} \cite{sdv-dev} to the unchanging context table. This allows us to understand any correlations between the context columns. We call this the \emph{context model}.
    \item Then we apply a Sequential Modeling algorithm, CPAR, to each of the sequences. CPAR models the sequential order while taking the context into account. We describe CPAR in detail in the next section.
\end{enumerate}

When creating synthetic data, we run through the same two steps:
\begin{enumerate}
    \item The context model is able to synthesize brand new contexts. This is important because users typically do not want the original context values to be leaked in the synthetic data. For example, if the sequences represent healthcare patients, the synthetic context creates new patients with different combinations of attributes.
    \item CPAR then incorporates the newly created context and generates a plausible ordered sequence for it based on its values. For example, we create a synthetic sequence based on the attributes of each synthetic patients.
\end{enumerate}

As a result, the overall synthetic data contains new sequences for new contexts.


\section{The CPAR Model}\label{cpar}

The goal of a sequential model is to capture inter-row dependencies and synthesize them. The sequential model we created is called the Conditional Probabilistic Auto-Regressive model (CPAR). It is designed to model multi-sequence data where each sequence contains an unchanging context.

\begin{figure}[H]
    \centering
    \includegraphics[width=0.6\linewidth]{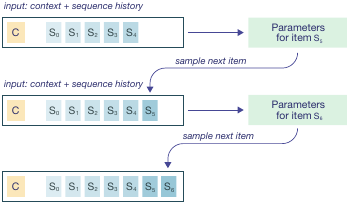}
    \caption{In this illustration, the unchanging context, $C$, is red while the sequence history is blue. If the current history includes $S_0, S_1, ..., S_4$, the model generates the distribution parameters for the next item $S_5$. We sample from the distribution to generate $S_5$ (dark blue), and it becomes part of the history. We repeat the process for the next item $S_6$ (darker blue).}
    \label{fig:cpar-summary}
\end{figure}

At a high level, CPAR expects as input the context along with the full history of the sequence (aka the inter-row dependencies). It then outputs the parameters needed to create the next row in that sequence. The new row becomes a part of the history and we repeat the process. This is illustrated in Figure \ref{fig:cpar-summary}.

In the rest of this section, we'll go into specific details of this algorithm.

\subsection{Neural Network Training}
CPAR is a neural network-based model. The training process estimates the parameters to create every item in every sequence (denoted as $\pi$). It then updates the network with the estimates. This is shown in pseudo-code below for a multi sequence dataset.

\begin{algorithm}
\caption{One training epoch}\label{alg:cap}
\begin{algorithmic}
\State Loss: $\mathcal{L}$
\State Neural Network \\

\For{Sequence $S^{(i)}$}
    \State $C^{(i)} \gets \text{Context}\left(S^{(i)}\right)$
    \For{Step $S^{(i)}_t$}
        \State $\pi^{(i)}_{(t,0)}, \pi^{(i)}_{(t,1)}, ... \gets \text{Neural Network}\left(C^{(i)}; S^{(i)}_0, S^{(i)}_1, ... S^{(i)}_{t-1}\right)$
        \State $\mathcal{L} \gets \mathcal{L} + \mathcal{L}\left(S^{(i)}_t; \pi^{(i)}_{(t,0)}, \pi^{(i)}_{(t,1)}, ...\right)$
    \EndFor
\EndFor \\

\State $\text{Neural Network} \gets \min\left(\mathcal{L}, \text{Neural Network}\right)$
\end{algorithmic}
\end{algorithm}

Note that sequences end when there is a $1$ in the stop column. The neural network learns to predict the probability of terminating each sequence $i$ at a given step $t$ by using an additional parameter, $\pi^{(i)}_{t,\tau}$.

\subsubsection{Loss Function}
The loss function quantifies how close the output parameters are to creating the real value. For the Conditional PAR Model, we wrote a custom loss function that factors in 3 levels of data:
\begin{itemize}
    \item $i$: All sequences $S^{(0)}, S^{(1)}, ... $
    \item $t$: All rows in each sequence $S^{(i)}_0, S^{(i)}_1, ... $
    \item $j$: All the parameters for each row $\pi^{(i)}_{(t, 0)}, \pi^{(i)}_{(t, 1)}, ..., \pi^{(i)}_{(t, k-1)}$
\end{itemize}

The overall loss function is the sum of all the individual losses:

\begin{equation} \mathcal{L} = \sum_i\sum_t\sum_{j=0}^{k-1} \mathcal{L}\left(S^{(i)}_t, \pi^{(i)}_{(t,j)}\right)
\end{equation}

\begin{figure}[H]
    \centering
    \includegraphics[width=0.5\linewidth]{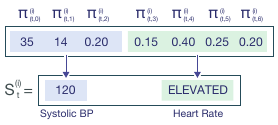}
    \caption{In this example, $S^{(i)}_t$ has two dimensions: Systolic BP (numerical) and Heart Rate (categorical). There are 7 parameters that are being used to estimate it: The first 3 parameters enable us to sample the Systolic BP (120) while the second 4 parameters enable us to sample the Heart Rate (ELEVATED).}
    \label{fig:cpar-parameters}
\end{figure}

Note that each row, $S^{(i)}_t$ is usually a multi-dimensional value. The parameters $\pi^{(i)}_{(t, 0)},\pi^{(i)}_{(t, 1)} ... $ describe the probability distribution of all its dimensions, although the number of parameters is different based on the data type. An example is illustrated in Figure \ref{fig:cpar-parameters}.

The number of parameters and loss function depends on the type of data.

\textbf{Continuous Numerical Data}
A continuous distribution is always described by 3 parameters: $\pi^{(i)}_{(t, \mu)}, \pi^{(i)}_{(t, \sigma)}, \text{ and } \pi^{(i)}_{(t, m)}$. Let's call these $\mu$,  $\sigma$ and $m$ for short. The $\mu$,  $\sigma$ represent the mean and standard deviation of a Gaussian distribution that would create the numerical value. The $m$ represents the probability that the value is missing.

Assume that the real value is $x$. Then the loss function is:

\begin{equation} \label{loss-numerical-cont}
\begin{split}
\mathcal{L}(x; \mu, \sigma, m) &= -\left(\log\left(f_{\mu, \sigma^2}\left(x\right)\right) + \log\left(1-m\right)\right): x \text{ is not missing} \\
\mathcal{L}(x; \mu, \sigma, m) & = -\log(m) : x \text{ is missing}
\end{split}
\end{equation}

Where $f$ is the probability density of a Gaussian curve with mean $\mu$ and variance $\sigma^2$. The $\log(1-m)$ and $\log(m)$ come from the Binary Cross-Entropy equation:

\[ H(p, q) = q\log(p) + (1-q)\log(1-p)\]

\textbf{Discrete Numerical Data}
A discrete distribution representing whole number counts is also described by 3 parameters: $\pi^{(i)}_{(t, r)}, \pi^{(i)}_{(t, \rho)}, \text{ and } \pi^{(i)}_{(t, m)}$. Let's call these $r$, $\rho$ and $m$ for short. The $r$ and $\rho$ represent the parameters of a Negative Binomial Distribution. The $m$ represents the probability that the value is missing.

The loss function is the same as the loss for continues values, except that the $f$ is now the probability density function of the Negative binomial distribution with parameters $r$ and $\rho$.

\begin{equation} \label{loss-numerical-disc}
\begin{split}
\mathcal{L}(x; r, \rho, m) &= -\left(\log\left(f_{r, \rho}\left(x\right)\right) + \log\left(1-m\right)\right) : x \text{ is not missing} \\
\mathcal{L}(x; r, \rho, m) & = -\log(m) : x \text{ is missing}
\end{split}
\end{equation}

\textbf{Categorical Data}
Categorical variables that contain a total of $N$ different categories are represented by $N$ parameters: $\pi^{(i)}_{(t,j)}, \pi^{(i)}_{(t, j+1)} \text{ and } \pi^{(i)}_{(t, j+N-1)}$, one for each category. Let's call these $\pi_0, \pi_1, ... \pi_{N-1}$ for short.

If each $j \in N$ represents a different category, then $\pi_j$ represents the probability that the value should be category $j$. This is similar to a one hot encoding scheme. If the real value for the category is $x$, then the loss is Cross-Entropy:

\begin{equation} \label{loss-categorical}
\begin{split}
\mathcal{L}(x; \pi_0, \pi_1, ... \pi_{N-1}) &= -\sum_{j \in N}x_j\log\left(\pi_j\right) \\ 
x_j &= 0 : j \text{ is not the correct category}\\
x_j &= 1 : j \text{ is the correct category}
\end{split}
\end{equation}

\textbf{Terminating a Sequence}
Since there is an extra parameter to signal the termination of a sequence, there is also a loss value associated with it.

This parameter is $\pi^{(0)}_{(t, \tau)}$, which we can call $\tau$ for short. Since $\tau$ is ultimately encoding for a binary variable, we can continue to use the same entropy-based loss function that we do for binary values.

\begin{equation} \label{loss-termination}
\begin{split}
\mathcal{L}(x; \tau) &=  -\left(x\log(\tau) + (1-x)\log(1-\tau)\right)\\ 
x &= 0 : \text{ the sequence has not terminated}\\
x &= 1 : \text{ the sequence has terminated}
\end{split}
\end{equation}

\subsubsection{Architecture}
The neural network architecture consists of a GRU \cite{GRU} in between two sets of dense layers. We use the Swish activation function \cite{Swish} in both dense layers. The most salient changes are the final layer, where we use different activation functions depending on the types of data:

\begin{itemize}
    \item For continuous numerical data, parameters $\mu$ and $\sigma$ use Softplus while parameter $m$ uses Sigmoid
    \item For discrete numerical data, parameter $r$ uses Softplus while $\rho$ uses the Sigmoid
    \item For categorical data, parameters $\pi_0, \pi_1, ... \pi_{N-1}$ together use the Softmax
\end{itemize}

\subsection{Neural Network Sampling}
When the training is complete, the neural network can be used for creating new sequences of synthetic data.

\subsubsection{Creating Sequences}

Let $S =  S_0, S_1 ... S_t$ be a single sequence of $t$ rows with an unchanging context, $C$. The trained Neural Network model estimates probability distribution parameters that are specific to creating the next sequence row, $S_{t+1}$:

\begin{equation} \label{sample-parameters}
\Pr(S_{t+1} | S_0, S_1, ... S_t; C)= f\left(S_{t+1}; \pi_{(t+1,0)}, \pi_{(t+1, 1)}, ..., \pi_{(t+1, k-1)}\right)
\end{equation}

There are a set number of parameters, $k$, that correspond to a probability density function of the next value, denoted as $f$. It is critical to estimate the distribution of the next row rather than the actual values: The synthetic data we want to create should cover a wide range of scenarios, including rare events. Models that estimate distributions are called \emph{Probabilistic Auto Regressive Models} (PAR).

To get an actual value, we can randomly sample from $f$. 

\begin{equation} \label{sample-data}
S_{t+1} \sim f
\end{equation}

The new value $S_{t+1}$ becomes part of the sequence. We can then repeat the process for the following item, $S_{t+2}$.

Note that in a multi sequence setting, we allow the users to specify the number of sequences to generate. By default, we generate the same number as the original data. The same overall process will be repeated for each different sequence, $S^{(i)}$. We can use the same neural network model for each $S^{(i)}$, because each sequence $S^{(i)}$ also has a different context $C^{(i)}$. The model is designed to condition on the context parameters, ensuring that each sequence has unique characteristics. This is why we call our model \emph{Conditional PAR} (CPAR).

\subsubsection{Interpreting the Parameters}
Each of the parameters $\pi_{(t+1,j)}$ has a different meaning. In order to sample step $S_{t+1}$ it's necessary to interpret the parameters based on the distribution they represent.

\textbf{Numerical Continuous Data.} Recall that this data is parameterized by $\mu$,  $\sigma$ and $m$. We can then sample a value from a Gaussian distribution with mean $\mu$ and variance $\sigma^2$. With probability $m$, we set it to be a missing value.

\textbf{Numerical Discrete Data.} Similarly, this data is parameterized by $r$, $\rho$ and $m$. We can then sample a value from a Negative Binomial Distribution with parameters $r$ and $\rho$. With probability $m$, we set it to be a missing value.

\textbf{Categorical Data}. This data is parameterized by $\pi_0, \pi_1, ..., \pi_{N-1}$, where each $j \in N$ represents a category, and each $\pi_j$ represents its probability. We can then select a category at random using the probabilities as weights.

\textbf{Terminating the Sequence}
To terminate a sequence, we use the generated parameter $\tau$. If $\tau > 0.5$, we end the sequence.

\section{Evaluation}\label{eval}

Our  objective is to compare the synthetic data created by a sequential model against the real data. We also benchmark it against synthetic data generated from a non-sequential model. This allows us to explain and quantify the effects of sequential modeling. We start by describing our experimental setup, including the datasets we use for evaluation. Finally, we report and discuss the the results\footnote{We intend to add more results in the future as we collect more datasets}.

\subsection{Experimental Setup}

In our experiment, we compare the CPAR model to the existing CTGAN \cite{ctgan} model. CTGAN is a popular GAN-based model that is available in the open source SDV ecosystem \cite{sdv-dev}. It is designed to learn the column shapes and correlations in tabular data, but it is not a sequential model. That is, it does not learn the concepts of multi-sequences or inter-row dependencies. 

\subsubsection{Data Generation Process}
Sequential data generation is straightforward.

\begin{enumerate}
    \item \textbf{Modeling.} We allow the model to learn a multi sequence dataset, inputting the sequence key, sequence index and context columns as parameters. We ran this model for 128 epochs.
    \item \textbf{Sampling.} We ask the model to generate the same number of sequences as the real data. The model will decide how many rows to generate for each sequence.
\end{enumerate}

With a non-sequential model, it possible to proxy sequential data generation in the following way:

\begin{enumerate}
    \item \textbf{Modeling.} We allow the model to learn the same multi sequence table. Since the model does not understand sequences or inter-row dependencies, it considers the sequence key as a categorical variable and the sequence index as a continuous value. We run this model for the same number of epochs.
    \item \textbf{Sampling.} We ask the model to generate the same amount of rows as the real data. We can then separate out sequences using the categorical sequence key and re-order the rows based on the sequence index.
\end{enumerate}

As a result, both models output a synthetic data table that contains multiple sequences with ordered rows.

\subsubsection{The Multi Sequence Dataset}

For evaluation, we used a publicly available multi-sequence dataset of NASDAQ stock prices \cite{sdv-dev}. Each company's stock is a different sequence. For each sequence, there is an unchanging context of Sector, Industry and MarketCap. Finally, there is a Date column that acts as the sequence index.

Table \ref{table:nasdaq-dataset} describes some other statistics about the table.

\begin{table}[H]
\centering
\begin{tabular}{|c | c |} 
 \hline
 Property & Count \\
 \hline
 Unique Sequences & 103 \\ 
 Total Rows & 25,784 \\
 Total Columns & 8 \\
 Context Columns & 3 \\
 Time-Varying Columns & 3 \\
 \hline
\end{tabular}
\caption{A summary of the multi sequence NASDAQ dataset.}
\label{table:nasdaq-dataset}
\end{table}

\subsection{Metrics}\label{metrics}

Our experimental setup yielded 3 different tables:
\begin{itemize}
    \item Real Data: The original dataset
    \item Synthetic Data (CPAR): Synthetic data generated from the sequential, CPAR model
    \item Synthetic Data (CTGAN): Synthetic data generated from the non-sequential, CTGAN model
\end{itemize}

In this section, we first define define an approach for measuring the similarity between real and synthetic data. Then we apply this approach on our dataset to report the results.

\subsubsection{Multi-Sequence Aggregate Similarity (MSAS)}

Since our data is both multi-column and multi-sequence, we cannot compute and compare a single statistic. Instead, we created an aggregation algorithm that compares multi-sequence data across columns for any statistic.  

Our basic algorithm computes a \textit{Multi-Sequence Aggregate Similarity} score, known as MSAS. The algorithm iterates through every column of every sequence and computes an underlying statistic $f$. For example, the length, mean, standard deviation, etc. We compare the two distributions for real and synthetic sequences using a 2-sample Kolmogorov–Smirnov (KS) test \cite{KSTest} and average the results per column. The details are shown in Algorithm \ref{alg:multi-seq-sim-metric}.

\begin{algorithm}
\caption{Multi-Sequence Aggregate Similarity (MSAS)}\label{alg:multi-seq-sim-metric}
\begin{algorithmic}
\State $C \gets \emptyset$
\For{Column $c$}
    \State $ X, X' \gets \emptyset$ \\

    \For{Real Sequence $S^{(i)}$, Synthetic Sequence $S'^{(j)}$}
        \State $X \gets X \cup f(S^{(i)}, c)$
        \State $X' \gets X' \cup f(S'^{(j)}, c)$
    \EndFor \\
    
    \State $C \gets C \cup \left(1 - \text{KS}(X, X')\right)$
\EndFor \\

\State MSAS $\gets \text{avg}(C)$

\end{algorithmic}
\end{algorithm}

The final MSAS score is in the range $[0, 1]$, where a 1 is the best score, indicating that the real and synthetic sequences have the same distributions for the underlying statistic $f$. We applied the MSAS algorithm to a variety of statistics ($f$):

\begin{itemize}
    \item The length of the sequence
    \item The sequence distribution: The mean, median and standard deviation
    \item Inter-row dependencies in a sequence: The average difference between a value in row $n$ and the row exactly $x$ steps after it in the sequence.
\end{itemize}

\subsubsection{Results}\label{results}

We applied the MSAS algorithm on the statistics described in the previous section. Table \ref{table:results} shows a summary of the results.

\begin{table}[H]
\centering
\begin{tabular}{|c || c | c | } 
 \hline
  & \multicolumn{2}{ | c  |}{ MSAS Score }\\
 \hline
 Statistic & Synthetic Data (CPAR) & Synthetic Data (CTGAN) \\
 \hline
 Sequence Length & 0.486 & 0.311 \\
 Column Mean & 0.667 & 0.644 \\
 Column Median & 0.654 & 0.835 \\
 Column Standard Deviation & 0.282 & 0.162 \\
 Inter-Row Difference (rows $n, n+1$) & 0.684 & 0.673 \\
 Inter-Row Difference (rows $n, n+5$) & 0.712 & 0.707 \\
 Inter-Row Difference (average) & 0.729 & 0.725 \\
 \hline
\end{tabular}
\caption{A summary of running the MSAS algorithm to compare the real NASDAQ dataset with the synthetic NASDAQ datasets generated from both the CPAR and CTGAN models. The full breakdown of results is available in the appendix.}
\label{table:results}
\end{table}

Our results show that the CPAR model creates synthetic sequences that are reasonably similar to the real sequences. Most of the MSAS scores are in the range of 0.6-0.7, indicating general similarity. In particular, the inter-row metrics are generally the highest, which makes sense because CPAR is designed to model the inter-row dependencies of sequential data. As the rows are further apart, the score tends to improve. One explanation is that in the real data, rows that are further apart are noisier and it is easy for CPAR to recreate that noise.

Note that none of the CPAR scores are 0.8 or above, which would indicate the highest quality. One possible factor is the number of training epochs. In future, we can explore increasing the number of epochs and evaluate how these metrics change as a result.

When comparing the CPAR model to the CTGAN model, our most salient insight is that there are no significant differences in most of these scores. Particularly, the CTGAN performs about equally as well as CPAR for the inter-row statistics. Even though the CTGAN is not modeling inter-row dependencies, it is able to effectively learn correlations between the timestamp and the other columns. This means that the overall, sorted synthetic data has properties similar to sequential modeling.

\subsection{Discussion}\label{discussion}

If sequential and non-sequential models create synthetic data of similar quality, why use a sequential model?

It is important to note that sequential models learn more about the data and are therefore more flexible in use. Sequential models are able to learn 3 dimensions of data (rows, columns and sequences) while non-sequential models are only able to learn 2 dimensions (rows and columns). Considering this, it a success if the CPAR model achieves similar results to the CTGAN model. It means we are not trading off the extra flexibility for synthetic data quality.

Below, we describe some scenarios where a non-sequential, 2D model is insufficient, requiring the use of the more flexible, 3D sequential model.

\subsubsection{Anonymizing Data with New Sequences}

 In this scenario, a user wants to use synthetic data to protect the real data. For this to succeed, it is vital that the synthetic data anonymizes the identities of the real data.

A non-sequential model learns the sequences as categorical variables and is only capable of generating those variables. That is, the generated sequences have a direct, 1-1 correlation with the real sequence. This leaks the identities of the real sequences, which can be problematic if they represent sensitive data such as healthcare patients.

\begin{figure}[H]
    \centering
    \includegraphics[width=0.6\linewidth]{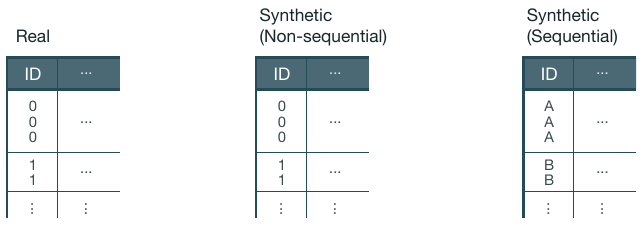}
    \caption{This example illustrates the synthetic data created from non-sequential vs. sequential models. In a non-sequential model, the synthetic data has the same sequence keys as the real data and the values have a direct correlation to real sequences. In a sequential model, even the sequences are fully synthetic. It's not just the names that are anonymized. Even the values cannot be traced back to a single, real sequence.}
    \label{fig:compare-seq-non-seq}
\end{figure}

However, our sequential model more powerful because it learns about sequences as a 3rd dimension. It
generates entirely new sequences do
not have a direct mapping to any single, real sequence. This has the effect of anonymizing the
identities of real sequences. This is illustrated in Figure \ref{fig:compare-seq-non-seq}.

\subsubsection{Augmenting the Original Data}
In another scenario, a user may want to generate many new sequences in order to augment the original data or create a much larger dataset.

As a corollary to the previous section, we note that a non-sequential model cannot generate more sequences than the real data contains. For example, if the real data contains $n$ unique sequences, the synthetic data can only create up to $n$ sequences.

By contrast, our sequential model can create any number of sequences. A user can take the real data with $n$ sequences and use the sequential model to generate more than $n$ new synthetic sequences for a richer overall dataset.

\subsubsection{Extrapolating Longer Sequences}
In this final scenario, a user wants to learn trends in the real data and use them to extrapolate synthetic data into the future. This can be used for forecasting.

A non-sequential model cannot effectively generate sequences that have a broader span than the original. If we use them to generate more synthetic rows, we will only succeed in filling out more data points in between the original range of observations. For example, if an original sequence contains steps observed between \texttt{Jan 2022} and \texttt{Jun 2022}, the model generally creates more rows inside this range. We cannot rely on a a non-sequential model to create broader sequences, such as forecasting future values beyond June.

Our sequential model is aware of step-wise, inter-row dependencies. Therefore, it can create sequences of any length, forecasting beyond the real data and into the future.


\section{Conclusion}

In this paper we described the Sequential SDV, a framework for generating synthetic sequential data as well as the CPAR model, a novel approach for modeling inter-row dependencies. We also described an algorithm, MSAS, for evaluating the quality of the synthetic data. Here, we provide some takeways.

\subsection{Key Findings}

\textbf{The Sequential SDV model can be applied to a variety of sequential datasets.} The model is specially designed to account for sequential data that may not have a time index, may have inconsistent intervals, may include multiple sequences in the same table, may have unchanging context columns and may have mixed data types. This shows the flexibility of the Sequential SDV framework in handling real world datasets.

\textbf{The CPAR model accounts for inter-row dependencies within a single sequence.} In our model, the neural network accepts the full sequence history as input, which means that it is capable of learning the dependency between any row $n$ and a subsequent row $n+x$.

\textbf{The CPAR model can generate different synthetic sequences because it conditions on the context and outputs parameters.} The neural network is designed to account for patterns in different sequences based on the context. Furthermore, it outputs distribution parameters instead of the next value. This means that sequence is not deterministic based on the context. It allows us to create multiple, different sequences, even if they have the same context. 

\textbf{The CPAR model learns high level information by learning multiple sequences as a third dimension of the data.} The CPAR learns broader trends about the sequences while treating them like a new dimension. This means that the CPAR model can create brand new sequences that do not directly correspond to any original sequence, it can create more sequences than the real data and it can extrapolate the data beyond the observed range. None of this is possible with a non-sequential model such as CTGAN.

\textbf{Despite the added complexity of modeling 3 dimensions, the CPAR model produces the same quality of synthetic data as regular, 2D tabular models.} Comparing the synthetic data from CPAR and CTGAN against the real data shows us that the quality of synthetic data is roughly equivalent for the two models. CPAR is able to perform as well as CTGAN even though it's learning more information about the data and capable of handling more use cases.

\pagebreak
\printbibliography

@inproceedings{
    sdv,
    author={N. {Patki} and R. {Wedge} and K. {Veeramachaneni}},
    booktitle={2016 IEEE International Conference on Data Science and Advanced Analytics (DSAA)},
    title={The Synthetic Data Vault},
    year={2016},
    volume={},
    number={},
    pages={399-410},
    doi={10.1109/DSAA.2016.49},
    ISSN={},
    month={10}
}

@inproceedings{ctgan,
  title={Modeling Tabular data using Conditional GAN},
  author={Xu, Lei and Skoularidou, Maria and Cuesta-Infante, Alfredo and Veeramachaneni, Kalyan},
  booktitle={Advances in Neural Information Processing Systems},
  year={2019}
}

@misc{sdv-dev,
  author = {DataCebo},
  title = {The Synthetic Data Vault},
  publisher = {GitHub},
  journal = {GitHub repository},
  howpublished = {\url{https://github.com/sdv-dev/SDV}},
}

@misc{GRU,
      title={Learning Phrase Representations using RNN Encoder-Decoder for Statistical Machine Translation}, 
      author={Kyunghyun Cho and Bart van Merrienboer and Caglar Gulcehre and Dzmitry Bahdanau and Fethi Bougares and Holger Schwenk and Yoshua Bengio},
      year={2014},
      eprint={1406.1078},
      archivePrefix={arXiv},
      primaryClass={cs.CL}
}

@misc{Swish,
      title={Searching for Activation Functions}, 
      author={Prajit Ramachandran and Barret Zoph and Quoc V. Le},
      year={2017},
      eprint={1710.05941},
      archivePrefix={arXiv},
      primaryClass={cs.NE}
}

@inproceedings{
    KSTest,
    author={F. J. Massey},
    booktitle={Journal of the American Statistical Association},
    title={The Kolmogorov-Smirnov test for goodness of fit},
    year={1951},
    volume={46},
    number={253},
    pages={68–78}
}
\pagebreak

\appendix

\appendixpage
\addappheadtotoc

\section{Experiment Results}

\subsection{Statistic Similarity}
\begin{table}[H]
\centering
\begin{tabular}{|c || c | c | c | c | c | c |} 
 \hline
  & \multicolumn{6}{ | c  |}{ MSAS Score }\\
 \cline{2-7}
 &  \multicolumn{3}{ | c  |}{Synthetic Data (CPAR)} & \multicolumn{3}{ | c  |}{Synthetic Data (CTGAN)} \\
 \cline{2-7}
 Statistic & Open & Close & Volume & Open & Close & Volume \\
 \hline
 Mean & 0.641 & 0.631 & 0.728 & 0.641 & 0.631 & 0.661 \\
 Median & 0.631 & 0.621 & 0.709 & 0.835 & 0.854 & 0.816 \\
 Standard Deviation & 0.155 & 0.214 & 0.476 & 0.049 & 0.058 & 0.379 \\
 \hline
\end{tabular}
\caption{MSAS scores for each of the 3 numerical columns in the table: Open, Close and Volume.}
\end{table}

\subsection{Inter-Row Similarities}
\begin{table}[H]
\centering
\begin{tabular}{|c || c | c | c | c | c | c |} 
 \hline
  & \multicolumn{6}{ | c  |}{ MSAS Score }\\
 \cline{2-7}
 &  \multicolumn{3}{ | c  |}{Synthetic Data (CPAR)} & \multicolumn{3}{ | c  |}{Synthetic Data (CTGAN)} \\
 \cline{2-7}
 Inter-Row Difference & Open & Close & Volume & Open & Close & Volume \\
 \hline
 1 & 0.604 & 0.641 & 0.806 & 0.592 & 0.585 & 0.842 \\
 2 & 0.619 & 0.653 & 0.822 & 0.604 & 0.607 & 0.853 \\
 3 & 0.629 & 0.655 & 0.822 & 0.614 & 0.614 & 0.856 \\
 4 & 0.640 & 0.662 & 0.824 & 0.625 & 0.621 & 0.862 \\
 5 & 0.645 & 0.668 & 0.824 & 0.628 & 0.629 & 0.863 \\
 6 & 0.643 & 0.672 & 0.832 & 0.637 & 0.634 & 0.866 \\
 7 & 0.654 & 0.673 & 0.831 & 0.640 & 0.637 & 0.865 \\
 8 & 0.657 & 0.678 & 0.835 & 0.644 & 0.645 & 0.865 \\
 9 & 0.662 & 0.679 & 0.839 & 0.648 & 0.649 & 0.869 \\
 10 & 0.663 & 0.685 & 0.835 & 0.657 & 0.652 & 0.867 \\
 11 & 0.665 & 0.685 & 0.839 & 0.659 & 0.658 & 0.867 \\
 12 & 0.671 & 0.689 & 0.842 & 0.661 & 0.661 & 0.868 \\
 13 & 0.674 & 0.692 & 0.840 & 0.662 & 0.657 & 0.869 \\
 14 & 0.675 & 0.693 & 0.838 & 0.665 & 0.667 & 0.869 \\
 15 & 0.678 & 0.700 & 0.836 & 0.667 & 0.665 & 0.870 \\
 16 & 0.679 & 0.697 & 0.840 & 0.669 & 0.669 & 0.868 \\
 17 & 0.681 & 0.699 & 0.839 & 0.669 & 0.673 & 0.868 \\
 18 & 0.681 & 0.698 & 0.841 & 0.679 & 0.676 & 0.869 \\
 19 & 0.683 & 0.703 & 0.838 & 0.677 & 0.677 & 0.868 \\
 20 & 0.687 & 0.705 & 0.844 & 0.678 & 0.681 & 0.872 \\
 21 & 0.688 & 0.706 & 0.844 & 0.681 & 0.680 & 0.869 \\
 22 & 0.687 & 0.704 & 0.844 & 0.679 & 0.681 & 0.867 \\
 23 & 0.694 & 0.707 & 0.841 & 0.684 & 0.684 & 0.866 \\
 24 & 0.694 & 0.707 & 0.842 & 0.684 & 0.687 & 0.864 \\
 25 & 0.695 & 0.707 & 0.842 & 0.683 & 0.686 & 0.866 \\
 \hline
\end{tabular}
\caption{MSAS scores for the inter-row differences between row $n$ and $n+x$ for each column, Open, Close and Volume. Here, $x$ varies from $1$ to $25$.}
\end{table}

\end{document}